\documentclass[conference]{IEEEtran}
\IEEEoverridecommandlockouts
\usepackage{cite}
\usepackage{amsmath,amssymb,amsfonts}
\usepackage{algorithmic,enumitem}
\usepackage{graphicx}
\usepackage{textcomp}
\usepackage{xcolor}
\usepackage{bm}
\usepackage[ruled,vlined,linesnumbered]{algorithm2e}

\def\BibTeX{{\rm B\kern-.05em{\sc i\kern-.025em b}\kern-.08em
    T\kern-.1667em\lower.7ex\hbox{E}\kern-.125emX}}

\usepackage{xcolor}
\newcommand\crule[4][black]{\textcolor{#1}{\rule[#4]{#2}{#3}}}
\definecolor{malecolor}{rgb}{0.1215,0.4666,0.705}
\definecolor{femalecolor}{rgb}{1.0,0.5,0.055}
\definecolor{firebrick}{rgb}{0.7, 0.13, 0.13}
\definecolor{darkgreen}{rgb}{0.0, 0.2, 0.13}
\definecolor{darkblue}{rgb}{0.0, 0.0, 0.55}

\begin{document}

\title{Plausible Counterfactuals: Auditing Deep Learning Classifiers with Realistic Adversarial Examples}
\author{\IEEEauthorblockN{Alejandro Barredo-Arrieta}
\IEEEauthorblockA{TECNALIA, Basque Research and Technology Alliance (BRTA)\\
48160 Derio, Bizkaia, Spain\\
alejandro.barredo@tecnalia.com}
\and
\IEEEauthorblockN{Javier Del Ser}
\IEEEauthorblockA{University of the Basque Country (UPV/EHU) \\
481013 Bilbao, Bizkaia, Spain\\
javier.delser@ehu.eus}
}

\maketitle
\thispagestyle{empty}
\pagestyle{empty}

\begin{abstract}
The last decade has witnessed the proliferation of Deep Learning models in many applications, achieving unrivaled levels of predictive performance. Unfortunately, the black-box nature of Deep Learning models has posed unanswered questions about what they learn from data. Certain application scenarios have highlighted the importance of assessing the bounds under which Deep Learning models operate, a problem addressed by using assorted approaches aimed at audiences from different domains. However, as the focus of the application is placed more on non-expert users, it results mandatory to provide the means for him/her to trust the model, just like a human gets familiar with a system or process: by understanding the hypothetical circumstances under which it fails. This is indeed the angular stone for this research work: to undertake an adversarial analysis of a Deep Learning model. The proposed framework constructs counterfactual examples by ensuring their plausibility, e.g. there is a reasonable probability that a human could generate them without resorting to a computer program. Therefore, this work must be regarded as valuable auditing exercise of the usable bounds a certain model is constrained within, thereby allowing for a much greater understanding of the capabilities and pitfalls of a model used in a real application. To this end, a Generative Adversarial Network (GAN) and multi-objective heuristics are used to furnish a plausible attack to the audited model, efficiently trading between the confusion of this model, the intensity and plausibility of the generated counterfactual. Its utility is showcased within a human face classification task, unveiling the enormous potential of the proposed framework.
\end{abstract}


%%%%%%%%%%%%%%%%%%%%%%%%%%%%%%%%%%%%%%%%%%%%%%%%%%%%%%%%%%%%%%%%%%%%%%%%%%%%%%%%
\section{Introduction}

The ever-growing achievements of complex models relying on powerful learning algorithms, such as those utilized in Deep Learning, have lately started to go beyond their academic boundaries towards their massive deployment in a manifold of application scenarios. This advent has been particularly notable in agriculture \cite{kamilaris2018deep}, Transportation \cite{del2019bioinspired}, Industry 4.0 \cite{diez2019data}, and a myriad of mobile applications to help users with their daily lives \cite{lane2015can, alsheikh2016mobile, yuan2014droid}. However, the design of these modeling techniques is often driven by its performance (\textit{let the model perform to its best for the task at hand}), thereby overlooking the context in which the model, once trained, will be in use (correspondingly, \textit{let the model be usable and understandable by its user}). To overcome this issue, quantifying and communicating the possibilities, limitations and caveats of Deep Learning models should be made compulsory for its practicality in real application environments. Unfortunately, the user receiving information on the model might not be an expert on Artificial Intelligence, which ultimately gives rise to a lack of trustworthiness, and his/her eventual reluctance to adopt the designed model in complex application niches.

In this context, current approaches for quantifying the performance of Deep Learning models clashes with the notions an overall user could appraise. For this reason, strategies and methods aimed to bridge the gap between these two realms of knowledge have been profusely explored in recent times. Among them, there are studies that inspect the robustness of Deep Learning models by means of adversarial attacks \cite{goodfellow2014generative, arjovsky2017wasserstein}. However, the output of these studies is still far from being user-friendly, since their practical implications when done for certain applications cannot be inferred. The estimation of the output confidence of the model has been also addressed in several works \cite{papernot2018deep,pmlr-v48-gal16,1707.07013}. Once again, the result may not be understandable for users without a background in Statistics. Further methods for explaining Deep learning models include the visualization of the internal representation of the model \cite{zeiler2010deconvolutional, simonyan2013deep}, the measurement of the amount of attention placed by the model to each of the inputs \cite{bach2015pixel} for a certain test sample, and many more covered in recently contributed prospective overviews \cite{arrieta2019explainable}.

The work presented in this manuscript joins this vibrant research area aimed at making Deep Learning models more interpretable and ultimately, more usable in practice. To this end, we propose an adversarial strategy to produce clear counterfactual explanations of the limits of a Deep Learning classifier. However, we further impose that the generated adversarial samples are \emph{plausible}, i.e., changes made on the input to the model have an appearance of credibility and believability without any computer intervention. To ensure plausibility, the proposed method makes use of GANs (Generative Adversarial Networks) in order to learn the underlying probability distribution of each of the features needed to create examples of an objective distribution (e.g. realistic human faces, which serves as an exemplifying use case throughout the paper). In parallel, a Deep Learning convolutional network separates two unknown classes (probability distributions) that are pertaining to the prior distribution (correspondingly, male and female faces). Our method allows for a search among samples of the first distribution to find realistic counterfactuals close to a given test sample that could be misclassified by the model (namely, a face of a male being classified as a female). As a result, our framework makes the user of the model assess its limits with an adversarial analysis of the probability distribution learned by the model, yet maintaining a sufficient level of plausibility for the analysis to be understood by a non-expert user.

The rest of the paper is organized as follows: Section 2 brings an analysis of the background literature for the techniques used in our study, whereas Section 3 presents materials and methods embedded in the proposed framework. Section 4 presents and discusses experimental results and, finally, Section 5 ends the work with several concluding remarks and a prospect of future research lines rooted on our findings. 

\section{Background}

As anticipated in the introduction, the proposed framework resorts to GANs for producing a realistic counterfactual analysis of image classification models that rely on CNNs (Convolutional Neural Networks). Since the ultimate goal is to favor the understanding of the model classification boundaries by an average user, the framework falls within the XAI (Explainable Artificial Intelligence) umbrella. Consequently, in this section we briefly contextualize and revisit the state of the art of such research areas. 

Generative adversarial networks were first introduced by Goodfellow in \cite{goodfellow2014generative}, bringing the possibility of using neural networks (\emph{function approximators}) to become generators of a desired distribution. Since their inception, GANs have progressively achieved photo-realistic levels of resolution and quality when synthesizing images of human faces. In general, a GAN architecture consists of two data-based models, which are trained in a mini-max game: one of the players (models) minimizes its error (loss), whereas the other maximizes its gain. In such a setup, multiple models have flourished to date, each governed by its strengths and vulnerabilities \cite{hindupur2017gan}. Interestingly for the scope of this paper, some of these were conceived with the intention of finding the pitfalls of a certain model and the ways to hack it \cite{goodfellow2014generative, arjovsky2017wasserstein}. Other GAN approaches aim at generating samples of incredibly complex distributions like photo-realistic human faces \cite{zhang2017stackgan, wu2019gp}. The framework proposed here hybridizes these two approaches by extending what was introduced by \cite{he2019attgan}, to include the idea of performing an adversarial analysis of a third model.

As for the modeling choice, we adopt the powerful capabilities of CNN \cite{russakovsky2015imagenet,krizhevsky2012imagenet} to capture spatial correlations from image data. Convolutional layers upon which CNNs are built allow for a space-wise abstraction of the input that, when trained via gradient back-propagation methods, give rise to image classification models of the highest performance. Although CNNs have been under the spotlight for a long time, the availability of huge loads of data and the capability of processing them efficiently has given rise to stacked convolutional architectures that perform incredibly well, at the cost of more complex, less understandable model structures \cite{lipton2018mythos}. The more complex the model is, the harder is to pinpoint the reasons for its failures, which motivates studies as the one presented in what follows.

The third technological pillar on which this contribution resides is the recent trend around model explainability, collectively referred to as XAI. In fact, XAI has recently become mainstream in the realm of Artificial Intelligence, unchaining a flurry of different approaches. The recent survey in \cite{arrieta2019explainable} provides an extensive review of the current state of the art on this matter, from which we extract \cite{ribeiro2016should, bach2015pixel} as the ones most closely related to the framework proposed in this paper. First, the LIME tool introduced in \cite{ribeiro2016should} allows for a linearization of a certain model's internal activations when faced with a specific test sample. This attempts to close the gap between the complexity of a model to predict the output and what users can understand to assess their inner functioning. Under this same scope, LRP (Layer-wise Relevance Propagation) implemented in SHAP \cite{lundberg2017unified} helps a user discern where, in the input image, the model focuses towards making its prediction. Once again, the target is to show, in understandable terms (an image overlaid by a heatmap) the reasons why the model outputs its decision. More recently, in \cite{liu2019generative} a similar GAN architecture to the one later explained is used to generate images by optimizing for the failure of the discriminator when freezing a couple of target variables. This allows them to study the representativeness of those classes within the dataset that has fed the model. However, a core difference with respect to this work is that the scheme in \cite{liu2019generative} does not present a user with the plausible changes he/she may undertake to produce an effective counterexample, nor does it explain under what circumstances the user should impose their criterion over that of the model.

\section{Materials and Methods} \label{sec:MatMet}

This section describes the data and methods comprised in the proposed counterfactual generation architecture. To this end, first data and its variations will be explained, followed by a statement of the mini-max problem under consideration, as well as the methods devised for solving it efficiently.

\subsection{Materials}

Although the proposed framework could be extrapolated to any image classification task, explanations and experiments discussed throughout this work consider the so-called Celebrity dataset \cite{liu2015faceattributes}, which consists of $200\cdot 10^3$ celebrity face images with annotated facial attributes. This dataset is built by considering forty different facial attributes, from which thirteen are chosen to feed the GAN: \texttt{Bald}, \texttt{Bangs}, \texttt{Black Hair}, \texttt{Blonde Hair}, \texttt{Brown Hair}, \texttt{Bushy Eyebrows}, \texttt{Eyeglasses}, \texttt{Gender}, \texttt{Mouth Open}, \texttt{Moustache}, \texttt{No Beard}, \texttt{Pale Skin} and \texttt{Age}. For the sake of computational efficiency, the images in high definition contained in this dataset have been downsized to $(128, 128, 3)$, in which the first two numbers refer to the height and width in pixels of the scaled image, and the third number denotes the number of color channels. The downsizing procedure has been carried out by means of Tensorflow's built-in \emph{bicubic} resampling method. We adopt the original train, validation and test partitions established in \cite{liu2015faceattributes}, with sizes $182,000$, $637$ and $17,363$ images, respectively. Further information on how these partitions have been made can be found in this reference.

\subsection{Problem Statement and Methods}

\begin{figure}[!ht]
	\vspace{-1mm}
	\centering
	\includegraphics[width=\columnwidth]{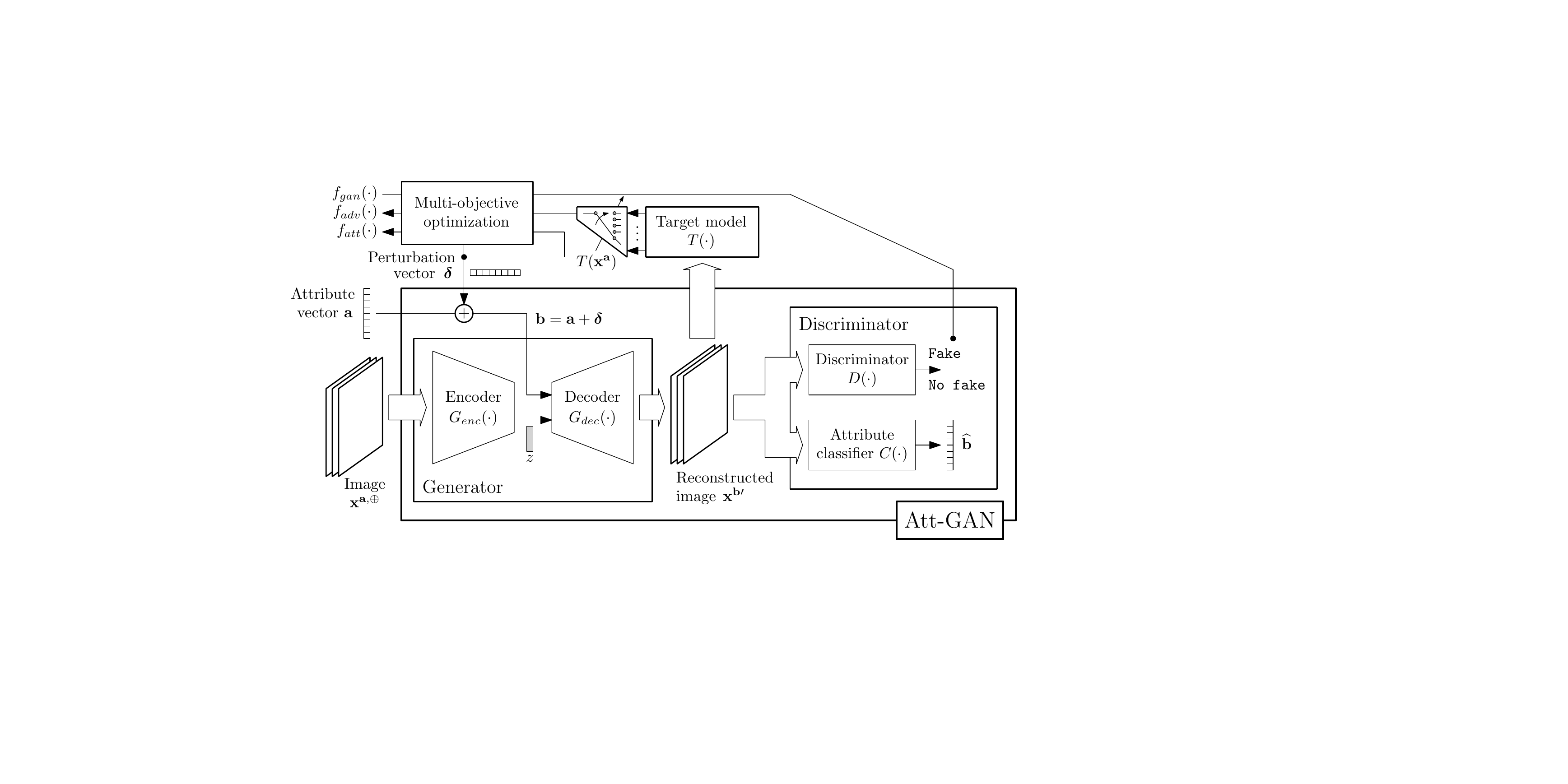}
	\caption{Block diagram of the proposed system comprising an AttGAN model and a multi-objective optimization model that infers the attribute modifications needed to produce plausible counterfactuals for target model $T(\cdot)$.}
	\label{fig:Architectures}
	\vspace{-1mm}
\end{figure}

The counterfactual generation framework proposed in this paper relies on a similar GAN architecture to the one introduced in \cite{goodfellow2014generative}. As shown in Figure \ref{fig:Architectures}, this architecture is composed by two models: one attempts to generate instances following certain characteristics (\emph{generator}), whereas the other determines whether the generated instance belongs to a distribution of interest (\emph{discriminator}). Originally the generator was fed with noise. More recently, other contributions \cite{arora2017theoretical, lyu2017auto} have replaced the generator model with an encoder-decoder structure to allow for a semi-supervised training, by which the generated content will pertain to the distribution of the input rather than noise. 

Several modifications have been imprinted to the conventional GAN structure to produce plausible counterfactuals of a third classification model. To begin with, following \cite{he2019attgan} we have added an extra input vector $\mathbf{b}$ to the decoder towards biasing the output of the generator. This new vector $\mathbf{b}$ can target a manifold of purposes; however, for the task at hand (human face recognition), the vector permits to tailor what attributes the generator changes in the input face in order to fool the discriminator. By devising and introducing an additional loss term $\mathcal{L}(\mathbf{b},\smash{\widehat{\mathbf{b}}^\prime})$ to the overall loss of the encoder-decoder GAN (with $\smash{\widehat{\mathbf{b}}^\prime}$ denoting the estimated vector of attributes of the reconstructed image by the decoder), we force the generator to maintain its input label at its output image, while changing certain facial attributes of the human face as per $\mathbf{b}$. Our addition to this architecture is non-intrusive: we include the classifier to be audited in parallel to the discriminator, yielding an additional loss term that quantifies the confidence of this classifier when producing its decision. 

At this point it is important to emphasize that the audited model is left aside the overall training process of the GAN for several reasons. To begin with, we assume that for practicality, the access to the audited model must be kept at its minimum (black-box analysis). Therefore, we only exploit the logits of the audited model, with no further information on its inner structure whatsoever. Furthermore, the goal of the discriminator is to decide whether the generated image follows the distribution of the training set, which must be regarded as a plausibility check. The task undertaken by the audited model can be assorted, for instance, to discriminate among \texttt{male} and \texttt{female}, \texttt{old} and \texttt{young} or any other classification problem alike. Therefore, the attribute vector $\mathbf{b}$ must balance between two objectives: 1) to maintain plausibility as per the discriminator, and 2) to confuse the audited model. A third objective can be also considered to account for the intensity of the changes inserted in the image as per $\mathbf{b}$: the subtler such changes are, the more unnoticed they will be in practice.

The above three goals can be formulated as a multi-objective optimization looking for a set of attribute vectors that produce, through the encoder-decoder of the GAN, images that achieve a Pareto-optimal balance between such goals. To recapitulate, one that determines if the discriminator is being fooled (reality check), another one that establishes whether the target fails to classify correctly the output of the GAN, and a third one that measures the intensity of attribute changes made to confuse the target model to be audited. 

To mathematically state this problem, let $\mathbf{x}^\mathbf{a}\sim P_{\mathbf{X}}(\mathbf{x})$ be an image on which the counterfactual analysis is to be made, which follows distribution $P_{\mathbf{X}}(\mathbf{x})$ and has attributes $\mathbf{a}$. We denote the generator as $G(\mathbf{x}^\mathbf{a},\mathbf{b})$, which is embodied by an encoder $G_{enc}(\mathbf{x}^\mathbf{a})$ and a decoder $G_{dec}(\mathbf{z},\mathbf{b})$, the latter receiving as its argument the compressed representation of the input image $\mathbf{z}=G_{enc}(\mathbf{x}^\mathbf{a})$, and an attribute vector $\mathbf{b}$. The image output by the generator is given by $\mathbf{x}^{\mathbf{b}\prime}= G_{dec}(G_{enc}(\mathbf{x}^\mathbf{a}),\mathbf{b})$. Ideally, $\mathbf{x}^{\mathbf{a}\prime} \approx \mathbf{x}^\mathbf{a}$, i.e. the reconstructed image $\mathbf{x}^{\mathbf{a}\prime}=G_{dec}(G_{enc}(\mathbf{x}^\mathbf{a}),\mathbf{a})$ should resemble $\mathbf{x}^\mathbf{a}$ itself.

A discriminator $D(\mathbf{x}^\prime)$ is placed next in the pipeline to determine i) whether the synthesized image $\mathbf{x}^\prime$ is visually realistic, and ii) whether the attributes of the generated images correspond to those reflected in $\mathbf{b}$. To this end, we learn an attribute classifier $C(\mathbf{x})$ from the annotated training image subset in order to compute an estimation $\smash{\widehat{\mathbf{b}}^\prime}$ of the attribute vector contained in its input image. 

With all these ingredients being defined, and following \cite{he2019attgan}, we can define the overall loss function that drives the learning algorithm of the encoder, decoder, discriminator and attribute classifier as a linear combination of the reconstruction and Wasserstein GAN losses, along with a loss reflecting that attributes $\mathbf{b}$ should be present in the reconstructed image. The training objective for encoder $G_{enc}(\mathbf{x}^\mathbf{a})$ and decoder $G_{dec}(\mathbf{z},\mathbf{b})$ are given by:
\begin{equation}\label{eq:minGG}
\min_{G_{enc},G_{dec}} \lambda_1 \mathcal{L}_{rec}(\mathbf{x}^{\mathbf{a}},\mathbf{x}^{\mathbf{a}\prime}) + \lambda_2 \mathcal{L}_{att}^G(\mathbf{b},\smash{\widehat{\mathbf{b}}^\prime}) + \mathcal{L}_{adv}^G(\mathbf{x}^{\mathbf{b}\prime}),
\end{equation}
where
\begin{align}
&\mathcal{L}_{rec}(\mathbf{x}^{\mathbf{a}},\mathbf{x}^{\mathbf{a}\prime}) = \mathbb{E}_{\mathbf{x}^\mathbf{a}\sim P_{\mathbf{X}}(\mathbf{x})} \left[\vert\vert\mathbf{x}^\mathbf{a}-\mathbf{x}^{\mathbf{b}\prime}\vert\vert_1\right],\\
&\mathcal{L}_{att}^G(\mathbf{b},\smash{\widehat{\mathbf{b}}^\prime}) = \mathbb{E}_{\mathbf{x}^\mathbf{a}\sim P_{\mathbf{X}}(\mathbf{x}),\mathbf{b}\sim P_{\mathbf{B}}(\mathbf{b})} \left[\sum_{n=1}^{N=|\mathbf{b}|} H(b_n,\widehat{b_n}^\prime)\right],\\
&\mathcal{L}_{adv}^G(\mathbf{x}^{\mathbf{b}\prime})= -\mathbb{E}_{\mathbf{x}^\mathbf{a}\sim P_{\mathbf{X}}(\mathbf{x}),P_{\mathbf{B}}(\mathbf{b})} \left[D(\mathbf{x}^{\mathbf{b}\prime})\right],
\end{align}
where $\mathbb{E}[\cdot]$ denotes expectation; $P_{\mathbf{B}}(\mathbf{b})$ indicates the distribution of possible attribute vectors in the range $\mathbb{R}[0,1]$; $H(b_n,\smash{\widehat{b_n}^\prime})$ is the cross-entropy of binary distributions given by $b_n$ and $\smash{\widehat{b_n^\prime}} \in \smash{\widehat{\mathbf{b}}^\prime} = C(\mathbf{x}^{\mathbf{b}^\prime})$; and $D(\mathbf{x}^{\mathbf{b}\prime})=0$ if $\mathbf{x}^{\mathbf{b}\prime}$ is predicted to be a fake. When placing the focus on the discriminator $D(\cdot)$ and the classifier $C(\cdot)$, the training goal is given by
\begin{equation} \label{eq:minDC}
\min_{D, C} \lambda_3 \mathcal{L}_{att}^C(\mathbf{x}^\mathbf{a},\mathbf{a})+\mathcal{L}_{adv}^D(\mathbf{x}^\mathbf{a},\mathbf{b}),
\end{equation}
with
\begin{align}
&\mathcal{L}_{att}^C(\mathbf{x}^\mathbf{a},\mathbf{a}) = \mathbb{E}_{\mathbf{x}^\mathbf{a}\sim P_{\mathbf{X}}(\mathbf{x})} \left[\sum_{n=1}^{|\mathbf{a}|} H(a_n,\smash{\widehat{a_n}}^\prime)\right],\\
& \mathcal{L}_{adv}^D(\mathbf{x}^\mathbf{a},\mathbf{b})\nonumber \\
&=-\mathbb{E}_{\mathbf{x}^\mathbf{a}\sim P_{\mathbf{X}}(\mathbf{x})} \left[D(\mathbf{x}^{\mathbf{a}})\right]+\mathbb{E}_{\mathbf{x}^\mathbf{a}\sim P_{\mathbf{X}}(\mathbf{x}),P_{\mathbf{B}}(\mathbf{b})} \left[D(\mathbf{x}^{\mathbf{b}\prime})\right],
\end{align}
where $\smash{\widehat{a_n}}^\prime \in C(\mathbf{x}^\mathbf{a})$, and coefficients $\{\lambda_i\}_{i=1}^3$ permit to balance the importance of the above losses during the training of the GAN architecture.

Once the GAN has been trained via back-propagation to meet the objectives in \eqref{eq:minGG} and \eqref{eq:minDC}, we exploit this trained generative architecture to find counterfactual examples for a given test sample $\mathbf{x}^{\mathbf{a},\oplus}$ and the model being audited, which we hereafter refer to as $T(\mathbf{x})$, with classes $\{\texttt{label}_1,\ldots,\texttt{label}_L\}$. Specifically, we seek a minimum perturbation to the attribute vector $\mathbf{a}$ (i.e. $\mathbf{b}=\mathbf{a}+\bm{\delta}$, with $\bm{\delta}\in\mathbb{R}^N$) that, through $G_{enc}$ and $G_{dec}$, yields a plausible image $\mathbf{x}^{\mathbf{b}\prime}$ that succeeds at confusing the target model $T(\cdot)$. The conflicting interplay between adversarial success, plausibility and intensity of the perturbation from which the counterfactual example is produced can be casted as a multi-objective optimization problem. Specifically,
\begin{equation}\label{Cost} 
\min_{\bm{\delta}\in\mathbb{R}^N}  f_{gan}(\mathbf{x}^{\mathbf{a},\oplus},\bm{\delta};G,D),f_{adv}(\mathbf{x}^{\mathbf{a},\oplus},\bm{\delta};G,T),f_{att}(\bm{\delta}), 
\end{equation}
where:
\begin{itemize}[leftmargin=*]
\item $f_{gan}(\mathbf{x}^{\mathbf{a},\oplus},\bm{\delta};G,D)$ quantifies the no-plausibility (\emph{unlikeliness}) of the generated counterfactual through $G$, which is given by the difference between the soft-max output of the discriminator $D(\dot)$ corresponding to $\mathbf{x}^{\mathbf{a},\oplus}$ and $\mathbf{x}^{\mathbf{b}\prime}$ (Wasserstein distance). The more negative this difference is, the more confident the discriminator is about the plausibility of the generated counterfactual $\mathbf{x}^{\mathbf{b}\prime}$; 
\item $f_{adv}(\mathbf{x}^{\mathbf{a},\oplus},\bm{\delta};G,T)$ is the probability that the generated counterfactual does not confuse the target model $T(\cdot)$, which is quantified by the negative value of the cross-entropy of the soft-max output of target model when queried with counterfactual $\mathbf{x}^{\mathbf{b}\prime}$; and
\item $f_{att}(\bm{\delta})$ gauges the intensity of adversarial changes made to the input image $\mathbf{x}^{\mathbf{a},\oplus}$, which is given by $\vert\vert \bm{\delta} \vert\vert_2$.
\end{itemize}

To efficiently find a set of attribute perturbations $\{\bm{\delta}\}$ balancing among the above three objectives in a Pareto-optimal fashion, we resort to multi-objective meta-heuristic optimization algorithms. Algorithm \ref{alg:counter} summarizes the process of generating counterfactuals for target model $T(\cdot)$, comprising both the training phase of the GAN architecture and the meta-heuristic search for counterfactuals subject to the three conflicting objectives stated above.
\begin{algorithm}[h!]
	\SetAlgoLined
	\DontPrintSemicolon
	\KwIn{Target model to be audited $T(\mathbf{x})$; GAN architecture ($G,D$); attribute classifier $C(\mathbf{x})$; annotated training set $\{\mathbf{x}^\mathbf{a}\}$; test image $\mathbf{x}^{\mathbf{a},\oplus}$ for producing the plausible counterfactual; weights $\{\lambda_i\}_{i=1}^3$}
	\KwOut{Set of plausible counterfactuals optimally balancing among $f_{gan}(\cdot)$, $f_{adv}(\cdot)$ and $f_{att}(\cdot)$}
	Train GAN architecture via back-propagation with the annotated training set and loss functions \eqref{eq:minGG} and \eqref{eq:minDC}\;
	Initialize a population of perturbation vectors $\bm{\delta}\in\mathbb{R}^N$\;
	\While{stopping criterion not met}{
		Apply search operators to the population to yield new perturbation vectors\;
		Evaluate their $f_{gan}(\cdot)$ (\emph{plausibility}), $f_{adv}(\cdot)$ (\emph{adversarial success}) and $f_{att}(\cdot)$ (\emph{intensity})\;
		Rank them in terms of Pareto optimality\;
		Retain the best perturbations in the population\;
	}
	Select non-dominated perturbations from population\;
	Produce counterfactual images by querying the GAN with $\mathbf{x}^{\mathbf{a},\oplus}$ and each selected perturbation vector
	\caption{Generation of plausible counterfactuals}
	\label{alg:counter}
\end{algorithm}

\section{Results and Discussion} \label{sec:Exp}

As has been mentioned previously, we exemplify the application of our proposed plausible counterfactual generation framework by designing an experimental setup on the Celebrity dataset. Specifically, we aim at auditing a target classifier $T(\cdot)$ trained with this dataset to distinguish among classes $\{\texttt{label}_1,\texttt{label}_2\}=\{\texttt{male},\texttt{female}\}$ ($N=2$), which is to be done by fabricating perturbations $\bm{\delta}$ of the attribute vector $\mathbf{a}$ of the test image $\mathbf{x}^{\mathbf{a},\oplus}$ that maintain the plausibility of the produced counterfactual image $\mathbf{x}^{\mathcal{b}\prime}$. Two different experiments have been designed, in which we alter the data to train the classifier:
\begin{enumerate}[leftmargin=*]
	\item We use the entire training set to create a classifier that discriminates between \texttt{Male} and \texttt{Female} (i.e. the target variable is \texttt{Gender}).
	\item We remove female examples with attribute \texttt{Blonde hair} from the training dataset, for reasons later disclosed throughout the discussion of the results.
\end{enumerate}

For the process of finding counterfactual examples in the above experiments, we employ the recently contributed jMetalPy framework for multi-objective optimization with meta-heuristic search algorithms \cite{benitez2019jmetalpy}. This framework eases the process of defining custom multi-criteria optimization problems, and provides off-the-shelf optimization algorithms for efficiently discovering approximation of the Pareto front of the defined problems. In order to select the best performing algorithm, we have performed a benchmark comparison of three of them: NSGAII (Non-dominated Sorting Genetic Algorithm, \cite{deb2002fast}), GDE3 (Generalized Differential Evolution, \cite{kukkonen2005gde3}) and SMPSO (Speed-constrained Multi-objective Particle Swarm Optimization, \cite{nebro2009smpso}). Hyper-parameters of these meta-heuristics have been tuned by means of an exhaustive search over a fine-grained grid of possible values, from where the best configuration was selected to be that yielding the best average hypervolume indicator value over 10 runs of every algorithm for $5$ images drawn at random from the test set. The reference point for the hypervolume was set to $(f_{gan},f_{adv},f_{att})=(0,-3,0)$. This benchmark concluded that the best algorithm is SMPSO with a swarm size equal to $100$ particles, polynomial mutation with probability $P_m=0.1$ and archive size equal to $100$ individuals. Further information on this comparison study is not disclosed for a better focus of forthcoming explanations.

The target classifier $T(\cdot)$ consists of a deep convolutional network that combines five sequential convolutional layers with batch normalization and leaky ReLu activations (rectified linear units with a 0.1 value under zero). This stacked set of convolutional layers end in a dense layer with a ReLu activation, followed by a second dense layer of a single unite activated by a sigmoid function that outputs the probability of being of the target class. This model has been trained separately, and added to the overall architecture once the weights are loaded from the solution published in \cite{he2019attgan}, namely, a fully-functional, pre-trained attribute-based GAN able to generate facial attributes as desired. Some modifications have been made to the architecture and the loss functions to incorporate the target classifier to be audited. Therefore, the GAN training step of Algorithm \ref{alg:counter} (line 1) has not been needed to produce the results discussed in the following subsections.

\subsection{Experiment 1: Target Classifier trained with Entire Data}

\begin{figure}[!h]
	\vspace{-5mm}
	\centering
	\includegraphics[width=\columnwidth]{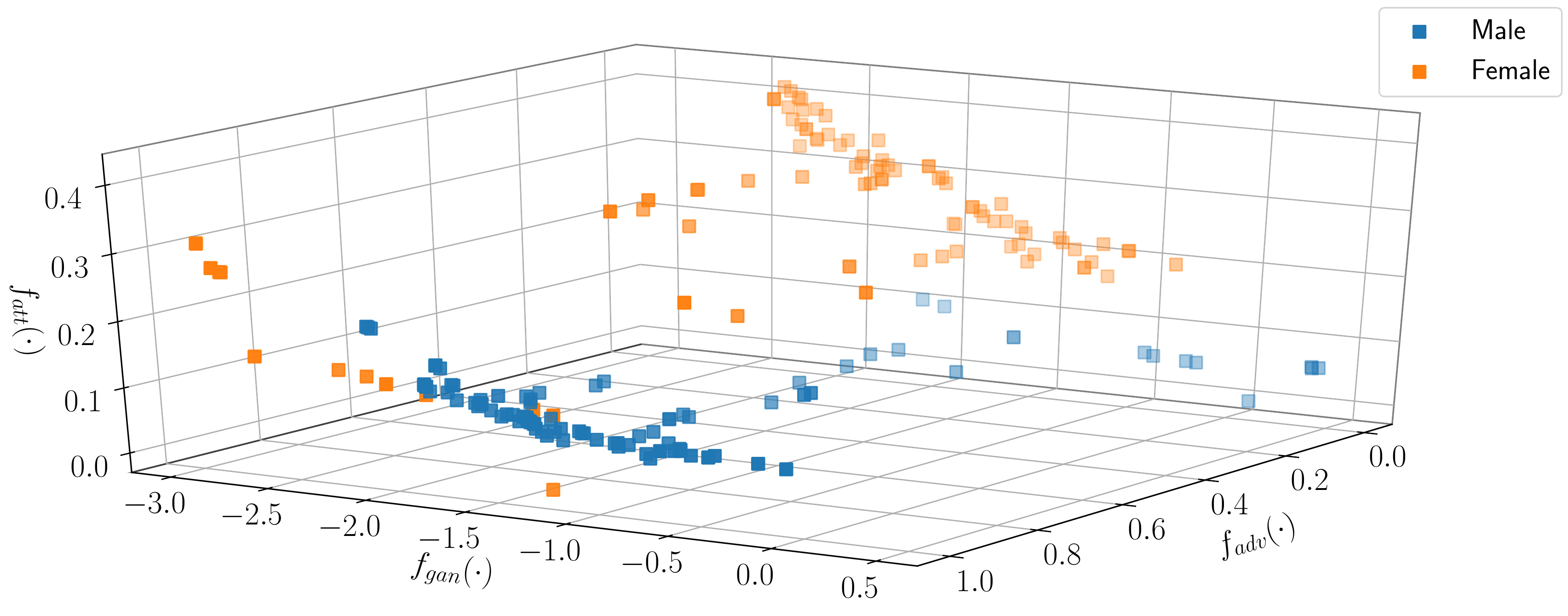} \\
	(a)
	\includegraphics[width=\columnwidth]{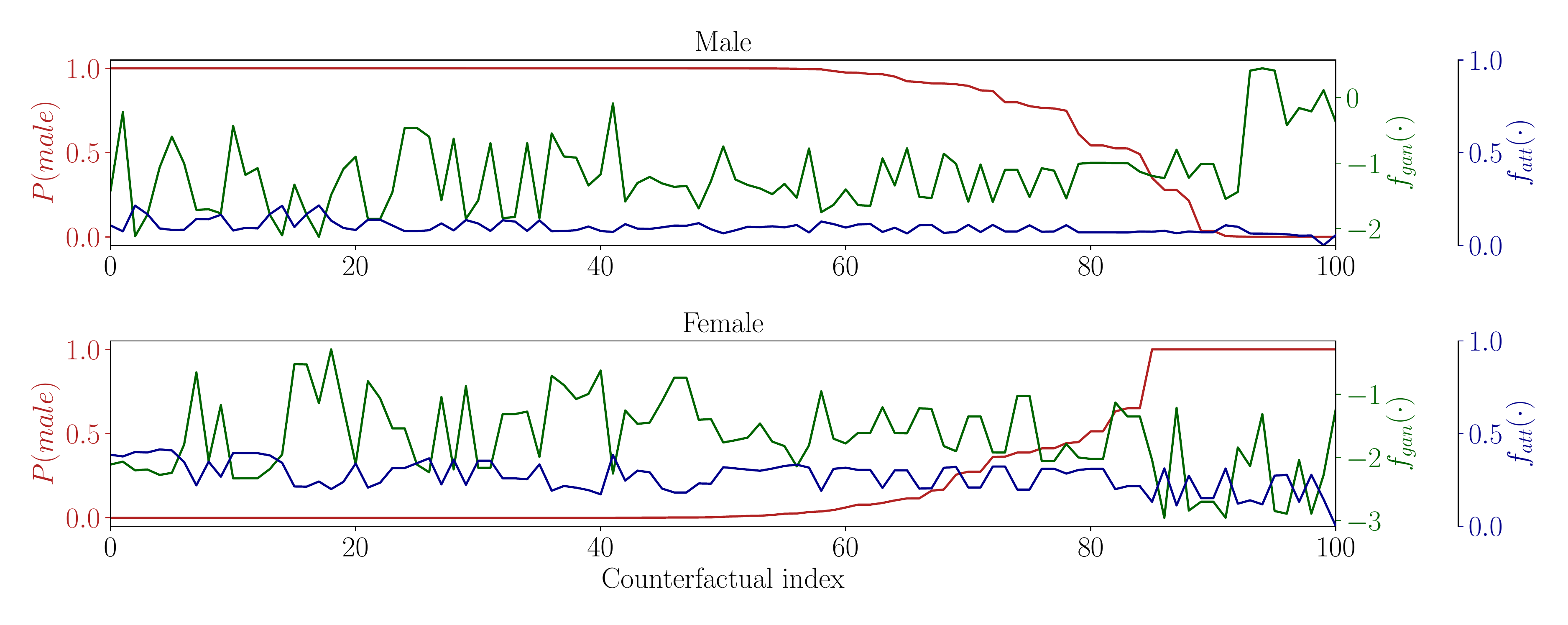} \\
	(b)
	\caption{(a) Estimated Pareto front for the first experiment, discriminating among counterfactuals corresponding to \texttt{Male} (\crule[malecolor]{0.15cm}{0.15cm}{0pt}) and $\texttt{Female}$ (\crule[femalecolor]{0.15cm}{0.15cm}{0pt}) test image; (b) probability of being classified as \texttt{male} by $T(\cdot)$ (\crule[firebrick]{0.25cm}{0.03cm}{0.6mm}), value of $f_{gan}(\cdot)$ (\crule[darkgreen]{0.25cm}{0.03cm}{0.6mm}) and $f_{att}(\cdot)$ (\crule[darkblue]{0.25cm}{0.03cm}{0.6mm}) for the whole set of counterfactuals contained in the estimated Pareto front of (a).}
	\label{fig:Exp1Graph}
\end{figure}

\begin{figure*}[h]
	\centering
	\includegraphics[width=1.95\columnwidth]{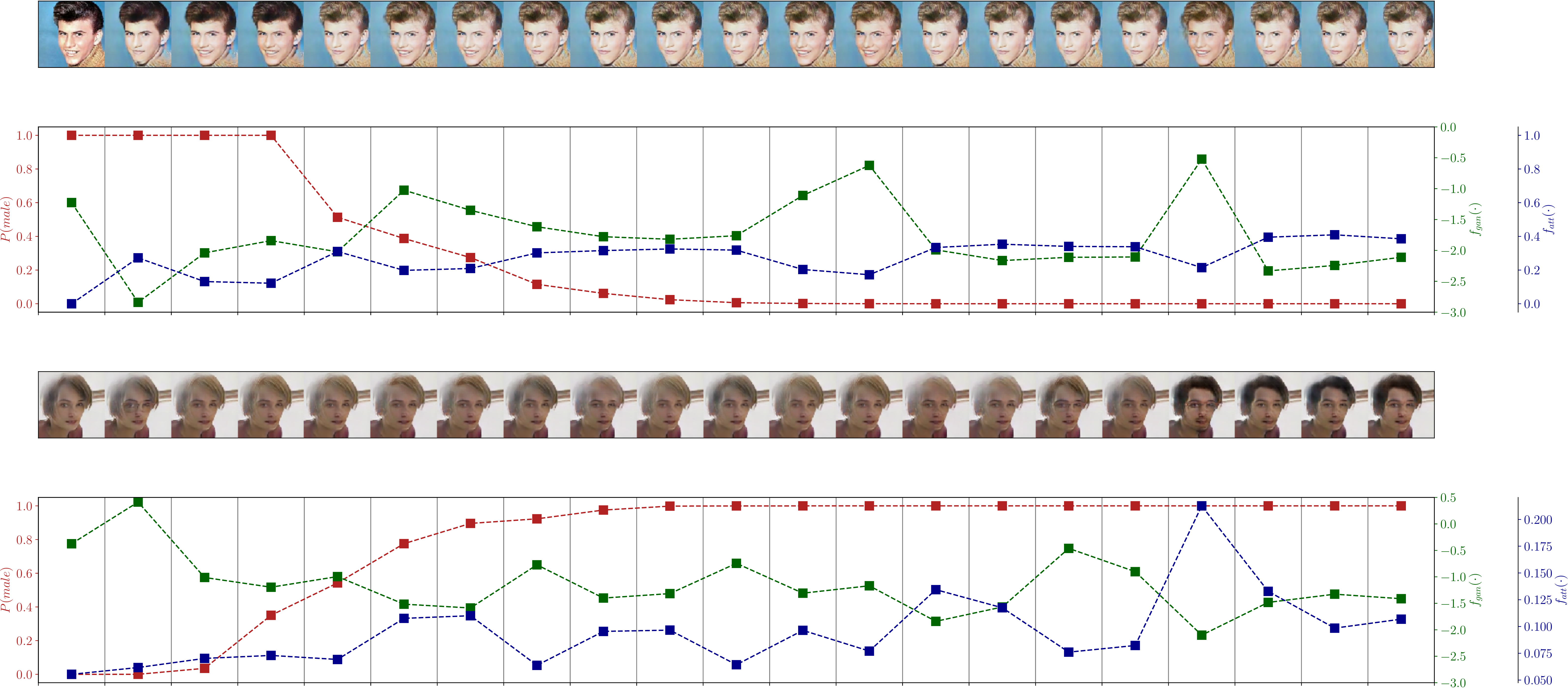}
	\caption{Two illustrative examples of Experiment 1: the original image $\mathbf{x}^{\mathbf{a},\oplus}$ (leftmost image), followed by counterfactuals belonging to the estimated Pareto front, thus balancing differently among plausibility, intensity and adversarial success. In the case of the male image, the insertion of blonde hair is decisive to confuse the audit classifier $T(\cdot)$, which rises suspicion about a potential bias of $T(\cdot)$ to concentrate on the hair color to produce its prediction.}
	\vspace{-3mm}
	\label{fig:Exp1Img}
\end{figure*}

Figures \ref{fig:Exp1Graph} and \ref{fig:Exp1Img} summarize the results obtained from the first experiment. On the one hand, Figure \ref{fig:Exp1Graph}.a depicts the Pareto front achieved for test images corresponding to male and female images, whereas Figure \ref{fig:Exp1Graph}.b shows the progression of the probability from the target classifier $T(\cdot)$ corresponding to the \texttt{male} class, overlaid by the values of the $f_{gan}(\cdot)$ and $f_{att}(\cdot)$ of every point in the estimated Pareto front of Figure \ref{fig:Exp1Graph}.a. Counterfactuals are arranged in decreasing (male images) and decreasing order (female images) of the probability of class $\texttt{male}$ output by $T(\cdot)$.

We comment on this first plot before proceeding further. To begin with, the intensity of changes to audit model $T(\cdot)$ when the test image $\mathbf{x}^{\mathbf{a},\oplus}$ corresponds to a female human are in general more intense (i.e. higher value of $f_{att}(\cdot)$). It is also interesting to see that, for both genders, a crossing point can be noted in Figure \ref{fig:Exp1Graph}.b between $P(male)$ and $f_{gan}(\cdot)$. As shown next, this point determines the best plausible counterfactual corresponding to each test image.

We proceed with the discussion of the results of the first experiment by inspecting Figure \ref{fig:Exp1Img} (next page), which illustrates the generated images $\mathbf{x}^{\mathbf{b}\prime}$ from the perturbation vectors of the Pareto front depicted in Figure \ref{fig:Exp1Graph}.a. For the sake of space, only every tenth of the counterfactuals sorted as per Figure \ref{fig:Exp1Graph}.b are shown, thereby unfolding a representative excerpt of the diversity of images synthesized by our proposed framework. It is surprisingly accurate and recurrently happening over many examples in the test set that the aforementioned crossing point between $P(male)$ and $f_{gan}(\cdot)$ corresponds to the most plausible counterfactual, as can be verified by simple visual inspection. Furthermore, a similar exploration of the counterfactuals produced by many other images unveils that blonde hair is recurrently added to male images so as to make them be classified as \texttt{female} by $T(\cdot)$, and vice versa: darkened hair color is imprinted to female images to make them be predicted as \texttt{male}. 

\subsection{Experiment 2: Target Classifier trained with Skewed Data}

\begin{figure*}[!ht]
	\centering
	\includegraphics[width=1.95\columnwidth]{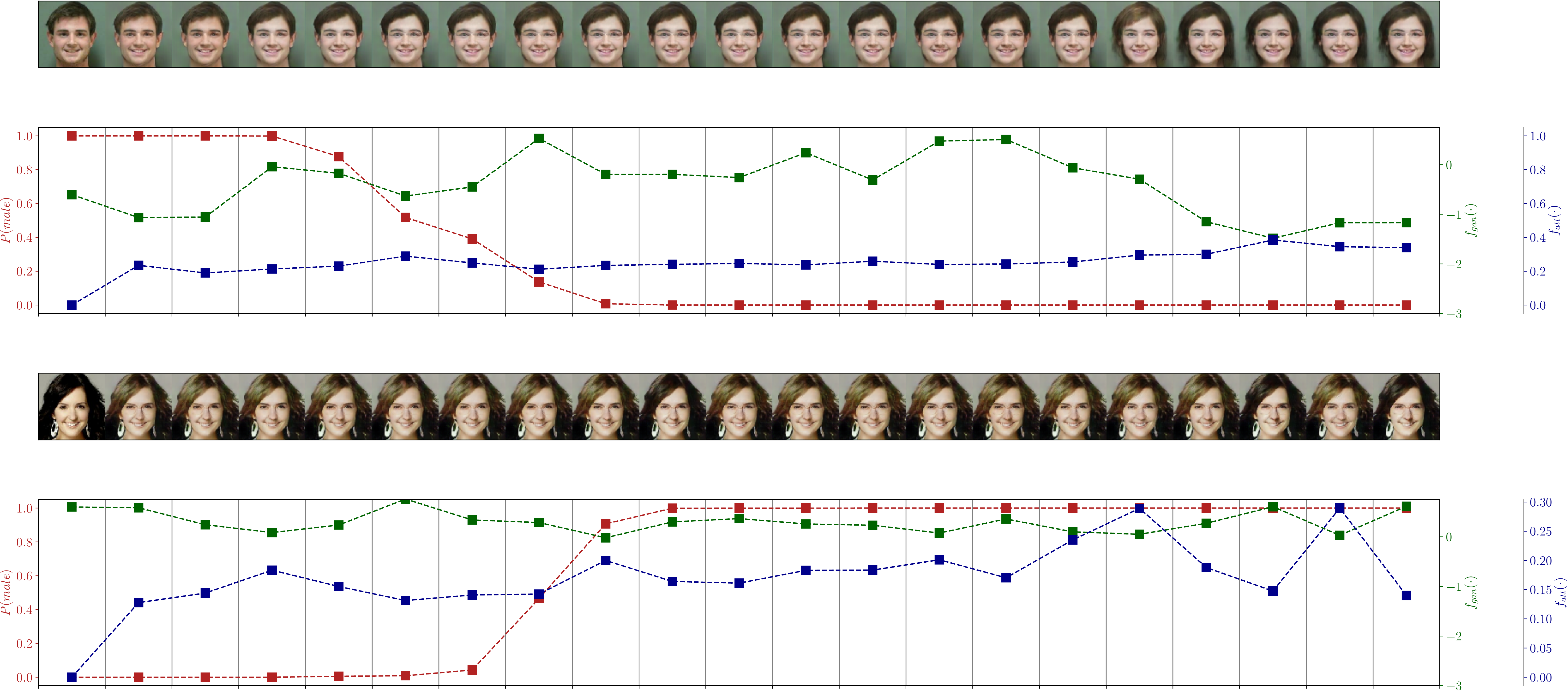}
	\caption{Two use cases generated under the conditions of Experiment 2 (no blonde females in the training set of the audit model), yielding plausible counterfactuals that are not affected by the excess of blonde female instances in the original dataset.}
	\label{fig:Exp2Img}
	\vspace{-3mm}
\end{figure*}

The previous experiment elucidates that the optimization process tends to turn the hair of female images to black, and the hair of male images to blond. There lies the rationale for devising a second experiment, where all images of female faces with blonde hair are removed from the training dataset of the target model. Figure \ref{fig:Exp2Img} shows two produced series of counterfactuals for this second experiment. In the case of the original male image (upper pair of plots), blonde haired counterfactuals are eventually produced, yet $P(male)$ plummets even if the hair is not blonde. A similar effect is perceived in the other depicted case, where the conversion of a female to a male realistically is attained by modifying other image attributes that are not necessarily related to the color of the hair (e.g. lower \texttt{bang}, higher \texttt{Age}) to increase $P(male)$. Although the color of the hair appears to still influence the production of counterfactuals for the female case, it is not as evident as in the case of Experiment 1. This corroborates the potential of our framework for discovering hidden sources of bias in classification problems that are not properly detected nor removed during the modeling phase.

\section{Conclusions and Future Research Lines} \label{sec:ConFut}

This work has elaborated on adversarial learning framework oriented towards the multi-objective synthesis of adversarial samples aimed at explaining the performance limits of Deep Learning classification problems. Unlike other adversarial learning studies wherein the goal is to produce subtle modifications to the input of the model, we aim at imprinting plausible changes that the user of the image could realistically imprint in the input to the model, so that plausibility is enforced to the produced counterfactual. To this end, we have focused on image classification, and hybridized an attribute GAN model \cite{he2019attgan} with a multi-objective meta-heuristic search engine to achieve counterfactuals for a given image that balance between three objectives related to the explainability of the classifier: \emph{plausibility}, \emph{adversarial success} and \emph{intensity} of the attribute perturbation. Two experiments have elicited interesting conclusions that go beyond the initially targeted goal of this study. Specifically, we have proven that the proposed framework may serve well to check underlying biasing phenomena present in a model and/or a dataset. Indeed, a subtle bias of the dataset (more blond women) was shown to be crucial for the adversarial success of male and female counterfactuals, informing the user about the risk of not tackling it (i.e. a simple change of color hair could eventually make a test image be classified wrongly by the audited model). 

In light of the global concern with the accountability of Artificial Intelligence methods, we plan to invest further efforts along several research directions rooted on the findings reported in this work. To cite a few, a closer look will be taken at how the counterfactual information generated by our framework can be fed back to the audited model and increase its robustness against adversarial or, alternatively, counteract the presence of bias in the training data. From the algorithmic side, Reinforcement Learning algorithms will be explored as an efficient alternative to multi-objective meta-heuristics, capable of learning the relationship between attribute changes and multi-criteria effect on the test image under consideration. 

\section*{Acknowledgments}

The authors would like to thank the Basque Government for its support through the EMAITEK and ELKARTEK funding programs. Javier Del Ser also receives support from the Consolidated Research Group MATHMODE (IT1294-19) granted by the Department of Education of the Basque Government.

\bibliographystyle{IEEEtran}
\bibliography{bib}

\begin{thebibliography}{10}
\providecommand{\url}[1]{#1}
\csname url@samestyle\endcsname
\providecommand{\newblock}{\relax}
\providecommand{\bibinfo}[2]{#2}
\providecommand{\BIBentrySTDinterwordspacing}{\spaceskip=0pt\relax}
\providecommand{\BIBentryALTinterwordstretchfactor}{4}
\providecommand{\BIBentryALTinterwordspacing}{\spaceskip=\fontdimen2\font plus
\BIBentryALTinterwordstretchfactor\fontdimen3\font minus
  \fontdimen4\font\relax}
\providecommand{\BIBforeignlanguage}[2]{{%
\expandafter\ifx\csname l@#1\endcsname\relax
\typeout{** WARNING: IEEEtran.bst: No hyphenation pattern has been}%
\typeout{** loaded for the language `#1'. Using the pattern for}%
\typeout{** the default language instead.}%
\else
\language=\csname l@#1\endcsname
\fi
#2}}
\providecommand{\BIBdecl}{\relax}
\BIBdecl

\bibitem{kamilaris2018deep}
A.~Kamilaris and F.~X. Prenafeta-Bold{\'u}, ``Deep learning in agriculture: A
  survey,'' \emph{Computers and Electronics in Agriculture}, vol. 147, pp.
  70--90, 2018.

\bibitem{del2019bioinspired}
J.~Del~Ser, E.~Osaba, J.~J. Sanchez-Medina, and I.~Fister, ``Bioinspired
  computational intelligence and transportation systems: a long road ahead,''
  \emph{IEEE Transactions on Intelligent Transportation Systems}, 2019.

\bibitem{diez2019data}
A.~Diez-Olivan, J.~Del~Ser, D.~Galar, and B.~Sierra, ``Data fusion and machine
  learning for industrial prognosis: Trends and perspectives towards industry
  4.0,'' \emph{Information Fusion}, vol.~50, pp. 92--111, 2019.

\bibitem{lane2015can}
N.~D. Lane and P.~Georgiev, ``Can deep learning revolutionize mobile sensing?''
  in \emph{16th ACM International Workshop on Mobile Computing Systems and
  Applications}, 2015, pp. 117--122.

\bibitem{alsheikh2016mobile}
M.~A. Alsheikh, D.~Niyato, S.~Lin, H.-P. Tan, and Z.~Han, ``{Mobile big data
  analytics using deep learning and Apache Spark},'' \emph{IEEE Network},
  vol.~30, no.~3, pp. 22--29, 2016.

\bibitem{yuan2014droid}
Z.~Yuan, Y.~Lu, Z.~Wang, and Y.~Xue, ``Droid-sec: deep learning in android
  malware detection,'' in \emph{ACM SIGCOMM Computer Communication Review},
  vol.~44, no.~4, 2014, pp. 371--372.

\bibitem{goodfellow2014generative}
I.~Goodfellow, J.~Pouget-Abadie, M.~Mirza, B.~Xu, D.~Warde-Farley, S.~Ozair,
  A.~Courville, and Y.~Bengio, ``Generative adversarial nets,'' in \emph{Adv.
  in Neural Information Processing Systems}, 2014, pp. 2672--2680.

\bibitem{arjovsky2017wasserstein}
M.~Arjovsky, S.~Chintala, and L.~Bottou, ``Wasserstein generative adversarial
  networks,'' in \emph{International Conference on Machine Learning}, 2017, pp.
  214--223.

\bibitem{papernot2018deep}
N.~Papernot and P.~McDaniel, ``Deep k-nearest neighbors: Towards confident,
  interpretable and robust deep learning,'' \emph{arXiv preprint
  arXiv:1803.04765}, 2018.

\bibitem{pmlr-v48-gal16}
Y.~Gal and Z.~Ghahramani, ``Dropout as a bayesian approximation: Representing
  model uncertainty in deep learning,'' in \emph{International Conference on
  Machine Learning}, vol.~48, 2016, pp. 1050--1059.

\bibitem{1707.07013}
A.~Subramanya, S.~Srinivas, and R.~V. Babu, ``Confidence estimation in deep
  neural networks via density modelling,'' \emph{arXiv preprint
  arXiv:1707.07013}, 2017.

\bibitem{zeiler2010deconvolutional}
M.~D. Zeiler, D.~Krishnan, G.~W. Taylor, and R.~Fergus, ``Deconvolutional
  networks,'' in \emph{{IEEE Conference on Computer Vision and Pattern
  Recognition}}, 2010, pp. 2528--2535.

\bibitem{simonyan2013deep}
K.~Simonyan, A.~Vedaldi, and A.~Zisserman, ``Deep inside convolutional
  networks: Visualising image classification models and saliency maps,''
  \emph{arXiv preprint arXiv:1312.6034}, 2013.

\bibitem{bach2015pixel}
S.~Bach, A.~Binder, G.~Montavon, F.~Klauschen, K.-R. M{\"u}ller, and W.~Samek,
  ``On pixel-wise explanations for non-linear classifier decisions by
  layer-wise relevance propagation,'' \emph{PloS one}, vol.~10, no.~7, p.
  e0130140, 2015.

\bibitem{arrieta2019explainable}
A.~{Barredo Arrieta}, N.~D{\'\i}az-Rodr{\'\i}guez, J.~Del~Ser, A.~Bennetot,
  S.~Tabik, A.~Barbado, S.~Garc{\'\i}a, S.~Gil-L{\'o}pez, D.~Molina,
  R.~Benjamins, and F.~Herrera, ``Explainable artificial intelligence {(XAI)}:
  Concepts, taxonomies, opportunities and challenges toward responsible {AI},''
  \emph{Information Fusion}, vol.~58, pp. 82--115, 2020.

\bibitem{hindupur2017gan}
A.~Hindupur, ``The {GAN} zoo: A list of all named {GANs},'' 2017.

\bibitem{zhang2017stackgan}
H.~Zhang, T.~Xu, H.~Li, S.~Zhang, X.~Wang, X.~Huang, and D.~N. Metaxas,
  ``Stackgan: Text to photo-realistic image synthesis with stacked generative
  adversarial networks,'' in \emph{IEEE International Conference on Computer
  Vision}, 2017, pp. 5907--5915.

\bibitem{wu2019gp}
H.~Wu, S.~Zheng, J.~Zhang, and K.~Huang, ``Gp-gan: Towards realistic
  high-resolution image blending,'' in \emph{ACM International Conference on
  Multimedia}, 2019, pp. 2487--2495.

\bibitem{he2019attgan}
Z.~He, W.~Zuo, M.~Kan, S.~Shan, and X.~Chen, ``Attgan: Facial attribute editing
  by only changing what you want,'' \emph{IEEE Transactions on Image
  Processing}, vol.~28, no.~11, pp. 5464--5478, 2019.

\bibitem{russakovsky2015imagenet}
O.~Russakovsky, J.~Deng, H.~Su, J.~Krause, S.~Satheesh, S.~Ma, Z.~Huang,
  A.~Karpathy, A.~Khosla, M.~Bernstein \emph{et~al.}, ``Imagenet large scale
  visual recognition challenge,'' \emph{International Journal of Computer
  Vision}, vol. 115, no.~3, pp. 211--252, 2015.

\bibitem{krizhevsky2012imagenet}
A.~Krizhevsky, I.~Sutskever, and G.~E. Hinton, ``Imagenet classification with
  deep convolutional neural networks,'' in \emph{Adv. in Neural Information
  Processing Systems}, 2012, pp. 1097--1105.

\bibitem{lipton2018mythos}
Z.~C. Lipton, ``The mythos of model interpretability,'' \emph{Queue}, vol.~16,
  no.~3, pp. 31--57, 2018.

\bibitem{ribeiro2016should}
M.~T. Ribeiro, S.~Singh, and C.~Guestrin, ``Why should i trust you?: Explaining
  the predictions of any classifier,'' in \emph{ACM SIGKDD International
  Conference on Knowledge Discovery and Data Mining}, 2016, pp. 1135--1144.

\bibitem{lundberg2017unified}
S.~M. Lundberg and S.-I. Lee, ``A unified approach to interpreting model
  predictions,'' in \emph{Adv. in Neural Information Processing Systems}, 2017,
  pp. 4765--4774.

\bibitem{liu2019generative}
S.~Liu, B.~Kailkhura, D.~Loveland, and Y.~Han, ``Generative counterfactual
  introspection for explainable deep learning,'' \emph{arXiv preprint
  arXiv:1907.03077}, 2019.

\bibitem{liu2015faceattributes}
Z.~Liu, P.~Luo, X.~Wang, and X.~Tang, ``Deep learning face attributes in the
  wild,'' in \emph{International Conference on Computer Vision}, 2015, pp.
  3730--3738.

\bibitem{arora2017theoretical}
S.~Arora, A.~Risteski, and Y.~Zhang, ``Theoretical limitations of
  encoder-decoder {GAN} architectures,'' \emph{arXiv preprint
  arXiv:1711.02651}, 2017.

\bibitem{lyu2017auto}
P.~Lyu, X.~Bai, C.~Yao, Z.~Zhu, T.~Huang, and W.~Liu, ``Auto-encoder guided
  {GAN} for chinese calligraphy synthesis,'' in \emph{IEEE International
  Conference on Document Analysis and Recognition}, vol.~1, 2017, pp.
  1095--1100.

\bibitem{benitez2019jmetalpy}
A.~Benitez-Hidalgo, A.~J. Nebro, J.~Garcia-Nieto, I.~Oregi, and J.~Del~Ser,
  ``{jMetalPy}: a {Python} framework for multi-objective optimization with
  metaheuristics,'' \emph{Swarm and Evolutionary Computation}, vol.~51, p.
  100598, 2019.

\bibitem{deb2002fast}
K.~Deb, A.~Pratap, S.~Agarwal, and T.~Meyarivan, ``{A fast and elitist
  multiobjective genetic algorithm: NSGA-II},'' \emph{IEEE Transactions on
  Evolutionary Computation}, vol.~6, no.~2, pp. 182--197, 2002.

\bibitem{kukkonen2005gde3}
S.~Kukkonen and J.~Lampinen, ``{GDE3: The third evolution step of generalized
  differential evolution},'' in \emph{IEEE Congress on Evolutionary
  Computation}, vol.~1, 2005, pp. 443--450.

\bibitem{nebro2009smpso}
A.~J. Nebro, J.~J. Durillo, J.~Garcia-Nieto, C.~C. Coello, F.~Luna, and
  E.~Alba, ``{SMPSO: A new PSO-based metaheuristic for multi-objective
  optimization},'' in \emph{IEEE Symposium on Computational Intelligence in
  Multi-Criteria Decision-Making}, 2009, pp. 66--73.

\end{thebibliography}

\end{document}